\documentclass[11pt,twocolumn]{article}

\usepackage{arxiv}

\usepackage[utf8]{inputenc} 
\usepackage[T1]{fontenc}    
\usepackage{hyperref}       
\usepackage{url}            
\usepackage{booktabs}       
\usepackage{amsfonts}       
\usepackage{nicefrac}       
\usepackage{microtype}      
\usepackage{amsmath}        
\usepackage{lipsum}
\usepackage{graphicx}
\graphicspath{ {./images/} }
\usepackage{amssymb}
\usepackage{float}
\usepackage{forest}
\usepackage{tikz}
\usetikzlibrary{arrows.meta}
\usepackage{multicol}
\usepackage{comment}
\usepackage{longtable, array, pdflscape}
\title{Compositional Policy Violations: When Step-Level Compliance Fails In Agentic AI Workflows}
\usepackage{bbding}
\usepackage{xcolor}
\usetikzlibrary{positioning,arrows.meta}
\usetikzlibrary{arrows.meta,positioning,calc,backgrounds,fit}
\usepackage{tabularx}
\usepackage{enumitem}
\usepackage{titlesec}
\usepackage{graphicx}
\usepackage{caption}
\usepackage{multirow}
\usepackage{xurl}

\makeatletter
\renewcommand{\@toptitlebar}{\hrule height 2\p@ \vskip 0.15in \vskip -\parskip}
\renewcommand{\@bottomtitlebar}{\vskip 0.18in \vskip -\parskip \hrule height 2\p@ \vskip 0.07in}
\renewcommand{\@maketitle}{%
  \vbox{%
    \hsize\textwidth
    \linewidth\hsize
    \vskip 0.05in
    \@toptitlebar
    \centering
    {\LARGE\sc \@title\par}
    \@bottomtitlebar
    \textsc{}\\
    \vskip 0.05in
    \def\And{%
      \end{tabular}\hfil\linebreak[0]\hfil%
      \begin{tabular}[t]{c}\bf\rule{\z@}{20\p@}\ignorespaces%
    }%
    \def\AND{%
      \end{tabular}\hfil\linebreak[4]\hfil%
      \begin{tabular}[t]{c}\bf\rule{\z@}{20\p@}\ignorespaces%
    }%
    \begin{tabular}[t]{c}\bf\rule{\z@}{20\p@}\@author\end{tabular}%
    \vskip 0.16in \@minus 0.05in \center{\today} \vskip 0.08in
  }
}
\makeatother

\makeatletter
\let\@oldthebibliography\thebibliography
\renewcommand{\thebibliography}[1]{%
  \@oldthebibliography{#1}%
  \setlength{\itemsep}{0pt plus 0.3pt}%
  \setlength{\parsep}{0pt}%
  \setlength{\topsep}{2pt}%
}
\makeatother

\author{
 Ashwini Kurady \\
  \texttt{ashwini@runctrl.ai} \\
   \And
 Sri Sai Charith Grandhi\\
  \texttt{charith@runctrl.ai} \\
  \And
 Rajesh Gupta \\
 \texttt{rajesh@runctrl.ai} \\
 \And
 Sumit Kumar \\
 \texttt{sumit.mamoria@gmail.com}
}

\titlespacing{\subsection}
{0pt}      
{4pt}      
{1pt}      

\usepackage{tikz}
\usetikzlibrary{shapes.arrows,positioning,calc}
\tikzset{
  stage/.style={
    rectangle,
    rounded corners=4pt,
    minimum width=3.5cm,
    minimum height=0.8cm,
    text centered,
    draw,
    line width=0.8pt,
    font=\footnotesize,
  },
  stage_gray/.style={stage, fill=gray!15, draw=gray!60},
  stage_blue/.style={stage, fill=blue!15, draw=blue!60},
  stage_purple/.style={stage, fill=purple!15, draw=purple!60},
  stage_coral/.style={stage, fill=orange!25, draw=orange!60},
  stage_red/.style={stage, fill=red!15, draw=red!60},
  arrow/.style={->,>=stealth,line width=0.8pt,gray!70},
}

\begin{document}

\makeatletter
\twocolumn[
  \begin{@twocolumnfalse}
  \maketitle
  \end{@twocolumnfalse}
]
\makeatother

\paragraph{Abstract - Agentic workflows now make consequential decisions in regulated settings, and the governance placed around them is almost entirely step-scoped: input-output classifiers, per-turn rails, and span-level evaluators. The policies organizations actually hold, such as referral thresholds, authority limits, and review requirements, are properties of the whole execution rather than of any one step. This mismatch admits a failure mode we call a \textbf{Compositional Policy Violation (CPV)}: every individual step passes its own check while the composed execution violates the governing policy. A predicate over a single step cannot evaluate a property that step does not determine, so no improvement in the accuracy of the step-scoped monitors detects this class. We define CPVs as the failure of step-level compliance to compose, and present a taxonomy of four types: Authority Creep, Threshold Laundering, Cumulative Sum Violation, and Context Collapse. We show that the correct repair for each class is dictated by where the guarded quantity mutates. We then introduce a provenance-aware runtime architecture that evaluates policies over complete execution traces, recomputing guarded quantities from raw provenance rather than the pipeline's derived representation.}

\section{Introduction}
Agentic workflows have moved into regulated decision making. American International Group (AIG), reporting on an early rollout built with Anthropic and Palantir, states that it compressed the timeline to review business by more than fivefold while raising data accuracy from 75\% to over 90\% \cite{anthropic2025financial}; Allianz \cite{allianz2026anthropic} and The Baldwin Group \cite{baldwin2026anthropic} have announced comparable deployments across underwriting operations. These are not laboratory demonstrations but production pipelines in which sequences of specialized agents ingest requests, enrich information, perform domain-specific analyses, and route decisions, with human oversight retained at designated approval or review steps. As a request moves through such a pipeline, it accumulates permissions, context, and prior determinations at each hop. The decision that reaches the final routing step is a function of everything the request picked up along the way, not just the state of any single step.

Governance for agentic workflows however, has converged on step-level predicates. Input-output classifiers evaluate prompts and responses in isolation\cite{inan2023llamaguard}. Programmable rails enforce constraints at turn level\cite{rebedea2023nemo}. Evaluators and guardrails attach at the span level\cite{langfuse2026datamodel,langsmith2026concepts}. Each mechanism asks the same question at the same grain: is this action permitted, given the state visible at this step?

The policies organizations actually hold are not of that shape. Referral thresholds, authority limits, and review requirements are properties of a trace (A trace is a complete ordered record of what a workflow did: every step, in sequence, with the state each step produced). Forcing them into step-level predicates creates an enforcement gap: a check that can reason only about the state in front of it cannot evaluate a property that depends on prior history or on what later steps will do. A workflow may therefore satisfy every individual check and still violate the policy it is subject to, with no component having failed.

We call this a Compositional Policy Violation (CPV). This is not a matter of imperfect calibration or insufficient monitor capability. It is structural: even perfectly accurate step-level monitors cannot detect this class, because the evidence needed to identify it exists only at the level of the composed execution. 

We note at the outset that trace-level monitoring is not itself novel; runtime verification and process compliance supply those techniques. Our contribution is to identify a specific governance failure mode in agentic workflows, characterize its forms, and operationalize its detection. 

The paper proceeds as follows. Section 2 situates our work relative to existing literature on guardrails, observability, evaluation, and human oversight. Section 3 defines Compositional Policy Violations (CPVs), differentiates them from conventional workflow defects, and introduces a four-part taxonomy: Authority Creep, Threshold Laundering, Cumulative Sum Violation, and Context Collapse. Section 4 shows that the correct repair (relocating a gate, adding one, or extending what an existing gate can see) is dictated by where the guarded quantity mutates. Section 5 presents a provenance-aware runtime architecture for CPV detection. We close by discussing future research in Section 6 and offering concluding remarks in Section 7.

\section{Related Work}

Existing approaches to AI governance, evaluation, and oversight largely assume that policy compliance can be determined at the boundary of individual steps. Modern guardrail systems such as Llama Guard, NeMo Guardrails, Constitutional Classifiers, and AgentSpec enforce constraints over prompts, responses, turns, or individual actions \cite{inan2023llamaguard,rebedea2023nemo,sharma2025constitutional,wang2025agentspec}. While effective for local violations, they do not evaluate whether a complete execution trajectory satisfies policies defined over the workflow.

Recent studies have shown that composition can defeat step-level safeguards in adversarial settings. Ahad et al.~\cite{ahad2026semantic} demonstrate that orchestrated subtasks can individually pass multiple safety classifiers while the combined plan violates security constraints; similar compositional failures have been observed across sessions, accumulated memory, and agent trajectories \cite{azarafrooz2026crosssession,mireshghallah2026cimemories,dhodapkar2026safetydrift}. However, these works assume an adversarial actor deliberately fragmenting behavior. In contrast, we study non-adversarial compositional failures where individually compliant actions, performed by independently scoped components, produce workflow-level policy violations through ordinary execution.

Observability and evaluation frameworks provide complementary capabilities but remain primarily step or outcome-oriented. Platforms such as Langfuse and LangSmith reconstruct execution traces through spans and runs but evaluate individual units rather than trace-level governance properties \cite{langfuse2026datamodel,langsmith2026concepts}. Similarly, process and outcome supervision approaches \cite{lightman2024verify} and agent failure taxonomies such as MAST \cite{cemri2025mast} analyze intermediate reasoning or failed executions, whereas CPVs represent workflows that complete successfully while violating policies defined over the accumulated trajectory.

Finally, human oversight remains the common fallback for high-risk AI workflows, yet studies show that reviewers often over-rely on automated recommendations and that oversight effectiveness decreases as automation increases \cite{green2022oversight,goddard2012automation}. These limitations motivate the need for runtime governance mechanisms that reason over complete execution histories. Our work addresses this gap through a provenance-aware architecture that combines policy evaluation, temporal reasoning, authority analysis, and sequence-level monitoring to detect compositional policy violations.

\section{Taxonomy}
\label{sec:taxonomy}

We now define this failure mode precisely. A Compositional Policy Violation (CPV) is a workflow in which every individual step satisfies the policy applied to it, yet the composed execution violates the policy governing the workflow as a whole. Three conditions hold together: each step is evaluated against a policy scoped to that step alone; each step performs its function correctly under that scoped policy; and the composed sequence violates a policy defined at the workflow level. No specification is breached and no component contains a bug. The defect lies in how the step-level specifications compose, which are locally sound but globally insufficient. 

This distinguishes a CPV from a conventional workflow defect. A race condition violates an invariant that assumed exclusive or ordered access to shared state, when concurrent execution interleaves that access unpredictably. A stale-state error reflects a step reading data past a validity window defined independently of any other step's timing. A missing recomputation is an omitted design element. A CPV, by contrast, occurs in purely sequential, correctly designed workflows with every local check in place. The violation is invisible to those checks not because monitoring is absent, but because the property being violated is defined over the trace, and no single step determines it.

\begin{figure}[H]
  \centering
 \includegraphics[width=1.0\columnwidth]{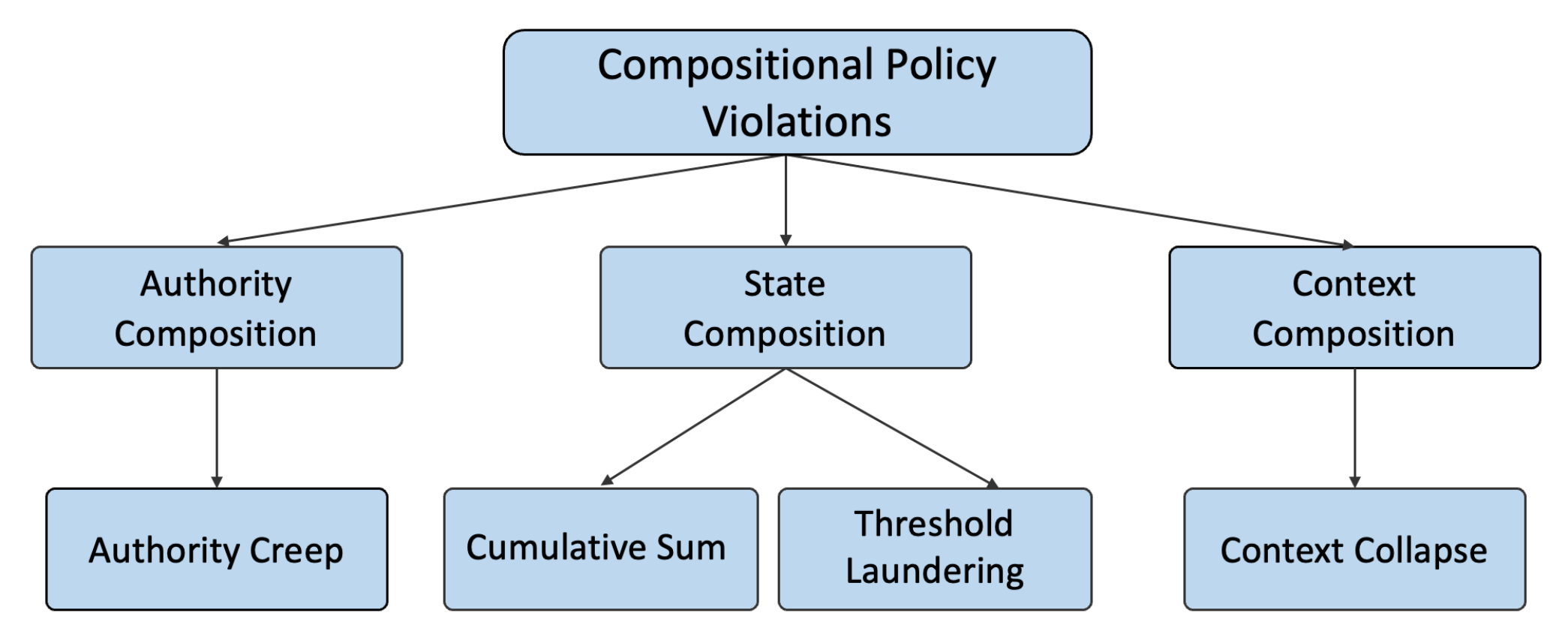}
  \caption{Taxonomy of compositional policy violations}
  \label{fig:cpv-taxonomy}
\end{figure}

We classify CPVs by the aspect of execution that becomes unsafe through composition rather than by a shared mechanism (Figure~\ref{fig:cpv-taxonomy}).
\textbf{Authority Creep} concerns who is entitled to decide, and arises when permissions accumulate across workflow steps. \textbf{Threshold Laundering} and
\textbf{Cumulative Sum Violation} concern an accumulated quantity: in the first, a gate checks the quantity correctly but a later step carries it past the limit with nothing re-checking; in the second, no gate exists at the aggregate scope, and individually compliant actions accumulate past the limit. \textbf{Context Collapse} concerns the representation on which the decision is made, and arises when each hand-off accurately summarizes what it received, yet the resulting record drifts further from the original submission with every step, until the final gate evaluates a case that no longer resembles the one the policy was written to address. The four share the CPV structure rather than a common mechanism: every local check passes, the composed execution violates the governing policy, and the evidence needed to see the violation is absent from every step. They differ in \emph{which} evidence is absent, and therefore in what a detector must retain. Section~\ref{sec:gateplacement} takes up the repair each class demands.

\paragraph{Terminology}
We use the following terms consistently throughout.

\begin{itemize}[leftmargin=1.4em,itemsep=1pt]
    \item \textbf{Predicate}: a boolean condition over a state or a trace.
    \item \textbf{Gate}: a workflow step that evaluates a predicate and routes
      execution on the result. Gates are part of the workflow and can change its
      behavior.
    \item \textbf{Monitor}: an observer that evaluates a predicate without
      altering execution. \emph{Step-level} when it is a function of the state at
      a single step; \emph{trace-level} when it is a function of the execution
      history.
    \item \textbf{Checkpoint}: the practice of evaluating one step or one change in isolation, without reference to the trace it belongs to.

\end{itemize}

\subsection{Authority Creep Violation}
A compositional policy violation in which a sequence of individually authorized agent operations transforms the representation of a guarded quantity such that an authority-routing gate(a rule that determines which principal - human tier, unit or escalation path is entitled to decide the case) evaluates to a different authority tier than the policy requires, even though the gate's own evaluation is satisfied. The workflow never claims decision authority. It shapes the input to the rule that allocates decision authority.

\begin{figure}[H]
  \centering
 \includegraphics[width=1.0\columnwidth]{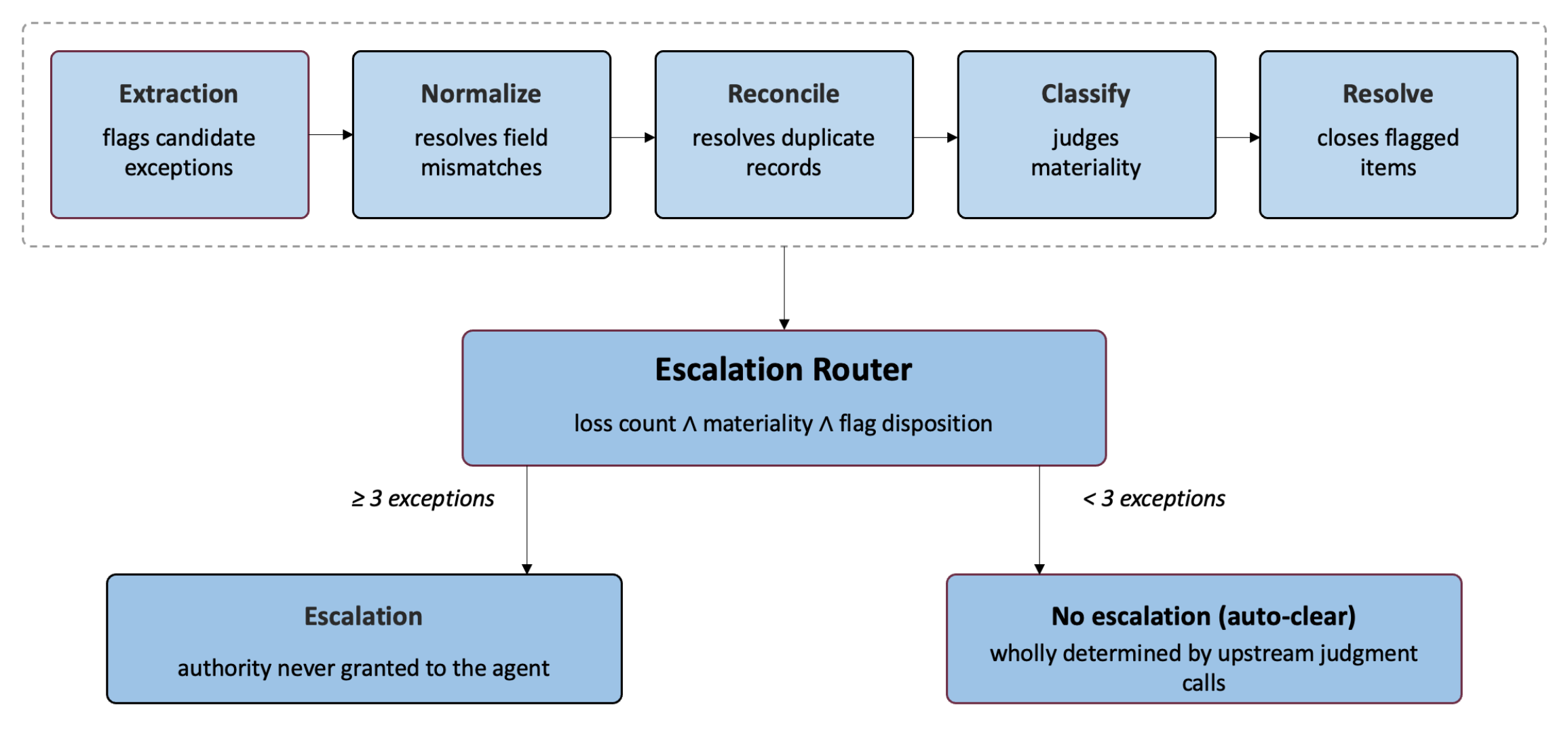}
  \caption{Authority Creep Violation in the sample under-writing pipeline. Expanding scope at each stage, illustrating how cumulative scope expansion emerges through composition rather than explicit authorization at any single step.}
  \label{fig:Authority_creep_example}
\end{figure}

Consider commercial insurance underwriting, shown in Figure~\ref{fig:Authority_creep_example}. The governing policy is that any account involving three or more material exceptions must be escalated. The pipeline holds five components. Extraction pulls submission data and flags candidate guideline exceptions; its authority is narrow, surfacing facts and deciding nothing about who reviews anything. Downstream of it sit components that normalize fields, reconcile duplicate records, classify materiality, and resolve flags. Each was scoped and approved on its own terms, against a requirement that existed before the others did, and none confers any decision authority over routing.

Specifically: normalizing COPE fields standardizes property attributes; reconciling loss runs manages duplicate records; classifying discrepancy materiality judges data quality; resolving flags closes out items the agent itself raised. That decision belongs to the escalation router alone, a fixed workflow step the agent is never granted control over.

Suppose extraction surfaces four candidate exceptions on a submission: an expired certificate, a COPE mismatch reporting two locations as one address, an unclassified prior loss, and a discrepant square-footage figure. Four exceptions clears the escalation threshold on its own. Normalization then resolves the COPE mismatch, the two locations are the same property under a corrected address leaving three, still at threshold. Loss-run reconciliation matches the unclassified prior loss to a claim already captured elsewhere in the submission's history and removes it as a duplicate, leaving two. Materiality classification determines the square-footage discrepancy falls within the policy's five-percent tolerance and is immaterial, leaving one. The escalation router reads a final count of one against a threshold of three, and does not escalate.

The distinction that matters here is between fixing an error and making a judgment call. Fixing an error means there was one correct answer and the step found it. For example, two database rows that are literally the same record, duplicated by a retry. Removing one is not Authority Creep, since no discretion over the guarded quantity was exercised. Making a judgment call means there were two or more reasonable interpretations, and the step picked one, and picking it made the exception disappear. For example, classifying the square-footage discrepancy as immaterial is one such call: the discrepancy could reasonably have been treated as material, and treating it as immaterial is what removed it from the count. This is the mechanism of the violation: each of several components, scattered across the pipeline, removes one exception on its own judgment call; only in composition do those separate calls add up to a decision that none of them was authorized to make. By the time the router evaluates the final count, the question it exists to answer(does this account need to be escalated?) has already been answered upstream, by components never granted authority to answer it.

\paragraph{Why per-step monitoring is blind to it:} Every transform stays inside its authorized scope and every step-level check passes. A monitor that sees one step and the policy scoped to it has no signal available, not a weak signal but none at all. The router itself is not defective either: it applies its threshold
correctly to the number it was handed. What no step holds is the relationship between the number the router reads and the evidence that number was derived from, and that relationship is the only place the violation is visible.

\paragraph{Trigerring Conditions:} An Authority Creep Violation is not an accident that can happen to any pipeline. It requires a specific arrangement of five conditions. Three of them describe the “Structural Preconditions” that makes the violation possible, and two Compositional Conditions that distinguish a genuine instance from ordinary misbehavior.

\paragraph{Structural Preconditions}
\begin{itemize}[leftmargin=1em]
\item \textbf{First}, the workflow must contain an authority-routing gate, as defined above — for example, escalation at three material exceptions, an SIU referral at two fraud indicators, or human-in-the-loop review above a risk band. Without such a rule there is no authority for the pipeline to creep past.
\item \textbf{Second}, one or more components must hold representation authority over that quantity's inputs: they are legitimately permitted to classify, score, merge, deduplicate, or suppress the very items the gate counts. This authority is granted for good reasons like noise reduction, precision, reviewer load.
\item \textbf{Third}, as established above, the gate reads the pipeline's transformed representation rather than recomputing from raw evidence under the policy's own definitions.
\end{itemize}
\paragraph{Compositional Conditions}
These conditions should be enforced as hard validity checks on every generated trace.
\begin{itemize}[leftmargin=1em]
\item \textbf{Local compliance must hold throughout}, as established above. If any step is individually non-compliant, the trace is contaminated. It contains an ordinary local violation that a conventional monitor would catch, and it tells us nothing about compositional governance.
\item \textbf{No single transform may be sufficient}: this condition concerns how the displacement is distributed across the pipeline: each transform, applied on its own to the raw evidence, must still route the case to the correct principal. This is also a no-culprit property because the identity of the responsible component depends on the execution order — determined by scheduling, not by anything the agent did. Consider two transforms in the opposite order. Consolidation first takes four exceptions to three, which still escalates; reclassification then takes three to two, which does not. Reverse them and reclassification is the innocent step while consolidation crosses the line. Same components, same scopes, same evidence, same outcome. Whichever transform runs second is the one that crosses the line, so the culprit is determined by scheduling rather than by anything any component did.
\end{itemize}

\subsection{Threshold Laundering}
\begin{table*}[t]
\centering
\small
\setlength{\tabcolsep}{6pt}
\renewcommand{\arraystretch}{1.25}
\begin{tabularx}{\textwidth}{@{}
>{\raggedright\arraybackslash}p{2.6cm}
>{\raggedright\arraybackslash}X
>{\raggedright\arraybackslash}X
>{\centering\arraybackslash}p{1.4cm}
>{\centering\arraybackslash}p{1.8cm}@{}}
\toprule
\textbf{Step} & \textbf{Action} & \textbf{Local check $P_i$} &
\textbf{State} & \textbf{Result} \\
\midrule
$s_1$\enspace Loss run intake
  & Read the submitted loss run and extract prior losses
  & Are losses correctly extracted and dated?
  & \$210K & Pass \\
$s_2$\enspace Referral gate
  & Apply the referral policy
  & $q(\sigma_2) \le \$250$K\,?
  & \$210K & Pass \\
$s_3$\enspace Supplemental intake
  & Process a later document carrying an additional \$75K closed claim
  & Is the claim correctly parsed and inside the five-year lookback window?
  & \$285K & Pass \\
\midrule
Quote (commit)
  & Commit to the auto-quote path at $\sigma_n$
  & ---
  & \$285K & $P(W) = \mathrm{fail}$ \\
\bottomrule
\end{tabularx}
\caption{Illustrative threshold laundering workflow. Every step passes its
local check, yet the committed state breaches the \$250K referral threshold
because nothing re-evaluates the gate after $s_3$.}
\label{tab:threshold-laundering}
\end{table*}

Threshold Laundering is a Compositional Policy Violation in which a quantity is checked against a limit at one point in the workflow, and a later step carries that quantity past that limit. No step misapplies the rule. The gate compares the quantity against the limit correctly, but does so before the quantity has finished changing. The value that gets committed is not the value that was checked, and nothing looks again. The rule is still sitting in the manual, fully in force, and it never fires.

Consider loss-history referral in commercial underwriting, as described in Table~\ref{tab:threshold-laundering}. Original policy: any account whose prior losses over the last five years exceed \$250K must be referred to an underwriting manager before it can be quoted. Submissions of this kind rarely arrive as a single document. Loss runs, supplemental carrier statements, and claim-closure notices are processed one after another - so the account's full loss picture builds up over the course of the workflow rather than being present at the start.
 
A loss run arrives first and is read as \$210K of prior losses. The referral gate checks this against the \$250K line, finds it under, and sends the account down the automatic-quote path. Its authority is narrow - it applies the rule to the number in front of it. It decides nothing about what arrives next. Later, a second document arrives carrying an additional \$75K closed claim, correctly dated inside the five-year window. The account now carries \$285K - over the line. But the referral gate has already run, and nothing re-checks the total. The account is quoted automatically, carrying \$285K of prior losses, with no manager referral.
 
Every step did its job. The ingestion agent read the loss run correctly. The referral agent applied the \$250K rule correctly to the number it was handed. The supplemental intake agent parsed a valid, in-window claim. None exceeded its authority, misreported a value, or omitted a required check. The policy was still broken.
 
 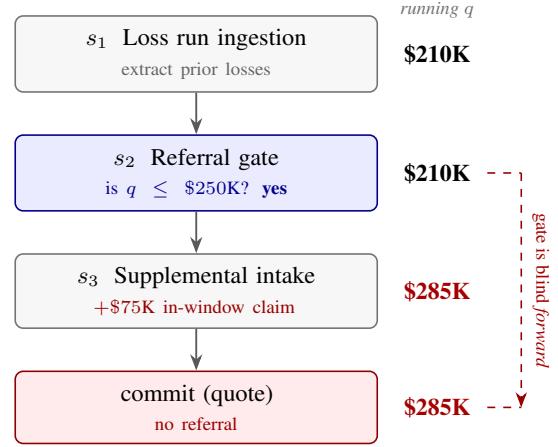
\begin{figure}[!t]
\centering
\begin{tikzpicture}[
  >=Stealth,
  node distance=0pt,
  box/.style={
    draw, rounded corners=3pt, line width=0.6pt,
    text width=\dimexpr0.55\columnwidth\relax,
    align=center, inner sep=5pt,
    font=\small,
  },
  pstep/.style   ={box, fill=gray!6,  draw=black!55},
  pgate/.style   ={box, fill=blue!8,  draw=blue!55!black},
  pcommit/.style ={box, fill=red!8,   draw=red!55!black},
  val/.style={font=\small\bfseries, anchor=west},
  flow/.style={->, line width=0.7pt, black!65},
]

\node[pstep] (s1) {
  $s_1$\; Loss run ingestion\\[1pt]
  {\scriptsize\color{black!60} extract prior losses}
};

\node[pgate, below=0.55cm of s1] (s2) {
  $s_2$\; Referral gate\\[1pt]
  {\scriptsize\color{blue!55!black} is $q \le \$250$K?\; \textbf{yes}}
};

\node[pstep, below=0.55cm of s2] (s3) {
  $s_3$\; Supplemental intake\\[1pt]
  {\scriptsize\color{red!60!black} $+\$75$K in-window claim}
};

\node[pcommit, below=0.55cm of s3] (s4) {
  commit (quote)\\[1pt]
  {\scriptsize\color{red!60!black} no referral}
};

\draw[flow] (s1) -- (s2);
\draw[flow] (s2) -- (s3);
\draw[flow] (s3) -- (s4);

\node[val, right=0.22cm of s1] (v1) {\$210K};
\node[val, right=0.22cm of s2] (v2) {\$210K};
\node[val, right=0.22cm of s3, text=red!65!black] (v3) {\$285K};
\node[val, right=0.22cm of s4, text=red!65!black] (v4) {\$285K};

\node[font=\scriptsize\itshape, text=black!55, above=0.10cm of v1] {running $q$};

\coordinate (a) at ($(v2.east)+(0.55,0)$);
\coordinate (b) at ($(v4.east)+(0.55,0)$);
\draw[dashed, line width=0.6pt, red!60!black] ($(v2.east)+(0.10,0)$) -- (a);
\draw[dashed, line width=0.6pt, red!60!black] ($(v4.east)+(0.10,0)$) -- (b);
\draw[->, dashed, line width=0.6pt, red!60!black] (a) -- (b);

\node[font=\scriptsize, text=red!65!black, rotate=-90,
      right=0.26cm of $(a)!0.5!(b)$, anchor=center]
  {gate is blind \emph{forward}};

\end{tikzpicture}
\caption{\textbf{Threshold Laundering.} The gate at $s_2$ applies the \$250K rule
correctly to the value in front of it. A later step adds a \$75K in-window claim,
so the committed value breaches the threshold, and no step re-evaluates the gate.}
\label{fig:threshold-laundering}
\end{figure}

This is a compositional policy violation because two checks make the difference, and both must hold:
\begin{itemize}[leftmargin=1em]
\item \textbf{Local compliance:} Every component stayed inside its own scope. The extractor extracted, the referral agent applied its threshold to the number it was handed, the intake agent parsed a claim. Every local check passes. No monitor watching individual steps sees anything at all. 
\item \textbf{No single culprit:} The gate alone, on the number it saw, is
correct: \$210K is under the line. The intake step alone is correct: it parsed a
valid \$75K claim, and escalation is not its job. Only the two composed carry the
account past the line with nobody watching.
\end{itemize}

\paragraph{Why per-step monitoring is blind to it.} The gate holds the rule but sees only the earlier \$210K - the \$75K has not arrived yet. The intake step sees the \$75K but holds no referral rule. So the one step that has the rule cannot see the final total, and the one step that creates the final total does not have the rule. Nobody ever holds both at once, and the breach slips through the seam between them. This is why no amount of tuning, calibration, or capability added to the monitor that only sees one step can catch the violation: the thing that needs to be seen is not present at any single
step.
 
One natural objection is that the workflow should simply run the gate last. For a single quantity this works. It does not generalise, and the reason is instructive. To re-check the account after the \$75K claim, the check has to know the running total from the earlier step \emph{and} apply the referral rule that belongs to the gate. A check that reaches back across steps and applies another step's rule is no longer a local check - it is a check over the whole account, which is exactly the trace-level check we argue for. And once a workflow gates several quantities that finalise at different steps - losses after intake, premium after rating, exposure after endorsements - there is no single position late enough to see all of them at once.
 
The violation is also not a failure of any agent's specification. Each agent met its specification exactly. The defect is a property of how the steps compose. It can only be repaired at that level: recompute the total on the committed account and re-ask the question before the account binds.

\subsection{Cumulative Sum Violation}

Threshold evasion through gate-bypass, as seen in Threshold Laundering, represents one compositional violation pattern. A related but distinct pattern emerges when no aggregation-level gate exists at all. A Cumulative Sum Violation (CSV) is a compositional policy violation in which individually compliant actions or signals collectively cross a policy threshold, such that the violation emerges only from their aggregate state and is not observable from any individual component in isolation.

For example, consider an AI purchasing agent authorized to buy office supplies on behalf of a company. The policy imposes two constraints: no individual purchase may exceed \$100 without approval, and total daily spending may not exceed \$250. Over the course of a day, the agent executes three purchases of \$90 each. Each purchase, when evaluated independently, satisfies the per-action constraint. No individual transaction violates the policy; the violation exists only in the composition of the transactions. A control mechanism that evaluates each action independently without maintaining aggregate state cannot detect this failure mode. This pattern has regulatory precedent in transaction structuring rules, where prohibited behavior is defined over a sequence of related transactions rather than any individual transaction in isolation~\cite{uscode5324}.

\begin{figure}[H]
  \centering
 \includegraphics[width=1.0\columnwidth]{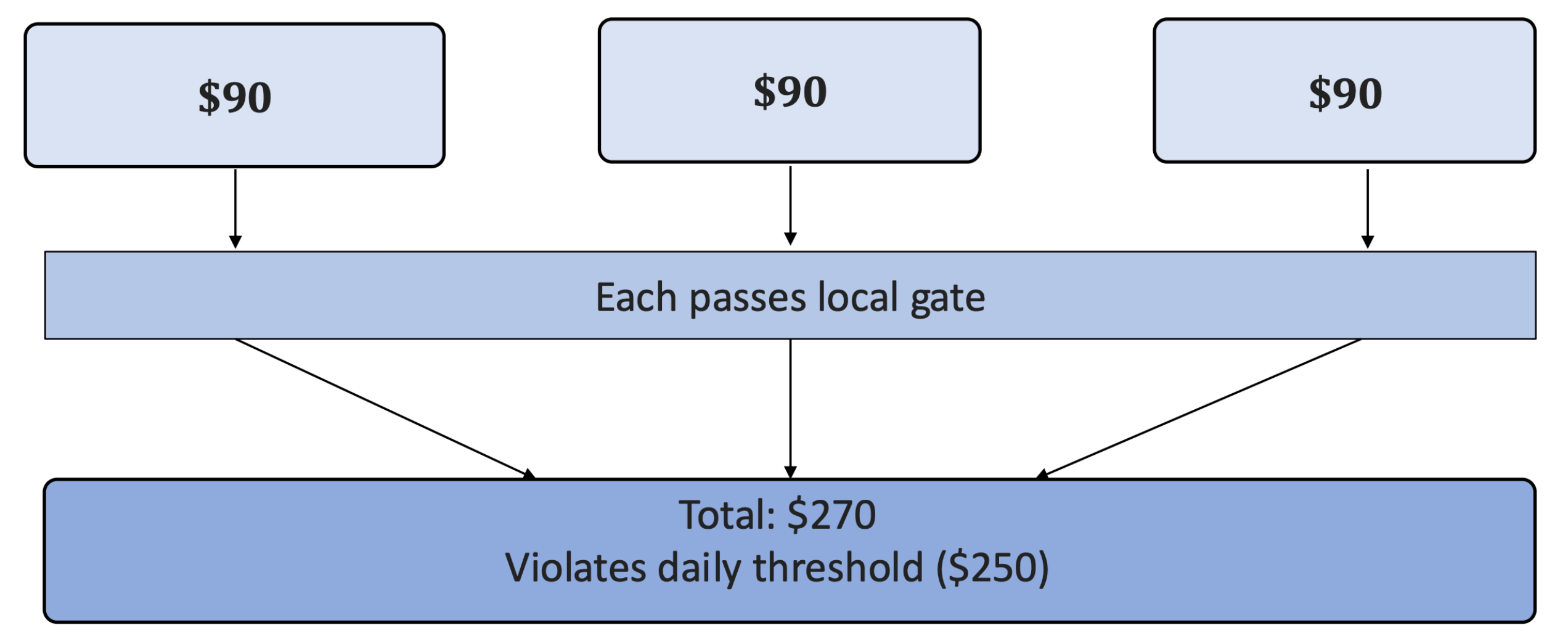}
  \caption{Cumulative Sum Violation: three compliant purchases aggregate to violate the daily spending threshold. Each purchase independently satisfies the per-action constraint (\$90 < \$100), but their composition exceeds the policy-level constraint (\$270 > \$250)}
  \label{fig:csv_purchasing_agent}
\end{figure}

The accumulated effect need not be a repeated sum of the same quantity. It may be a composite over heterogeneous signals: a loan application whose income verification, requested amount, and debt-to-income ratio each clear their own limit, while the composite credit-risk score they feed exceeds the referral threshold that no component computes.

\paragraph{Relationship to Threshold Laundering:} Both are threshold violations over accumulated state, but they differ in mechanism, in detection cost, and in repair. 

\emph{Mechanism:} In Threshold Laundering a gate for the guarded quantity exists and evaluates the policy correctly; the defect is its temporal position, since a later step revises the quantity it read. In a Cumulative Sum Violation, no gate at the aggregate scope exists at all; each component is gated at its own scope, and the composed quantity is never evaluated by anything.

\emph{Detection cost:} Once the committed state is available, detecting Threshold Laundering requires one recomputation of the guarded quantity and one comparison. Detecting a Cumulative Sum Violation requires maintaining a running aggregate over the whole trace, and where the aggregate spans entities, sessions, or invocations, correctly associating events with the entity the policy is defined over. The first is a question of \emph{when} the predicate is evaluated, the second, of \emph{what state must be carried} to evaluate it at all. 

\emph{Repair.} Threshold Laundering is repaired by re-evaluating an existing predicate on the committed state. A Cumulative Sum Violation is repaired by introducing a predicate that no step currently holds. A workflow can therefore exhibit one without the other: a correctly aggregated quantity gated too early is Threshold Laundering with no Cumulative Sum Violation, and an unaggregated quantity gated only at the level of individual actions is a Cumulative Sum Violation with no Threshold Laundering.

\subsection{Context Collapse Violation}
\label{sec:context-collapse}
\begin{table*}[t]
\centering
\footnotesize
\setlength{\tabcolsep}{6pt}
\renewcommand{\arraystretch}{1.1}
\begin{tabularx}{\textwidth}{@{}
>{\raggedright\arraybackslash}p{2.6cm}
>{\raggedright\arraybackslash}X
>{\raggedright\arraybackslash}X
>{\centering\arraybackslash}p{1.4cm}
>{\centering\arraybackslash}p{1.2cm}
>{\centering\arraybackslash}p{1.0cm}@{}}
\toprule
Step & Action & Local check & Facts kept & Hop drift & Result \\
\midrule
$s_1$ Intake          & Extract the submission package
                      & Facts extracted accurately?   & 1--5    & ---  & Pass \\
$s_2$ Enrichment      & Merge third-party data into a consolidated narrative
                      & Materially faithful to input? & 1,3,4,5 & 0.06 & Pass \\
$s_3$ Risk profiling  & Condense to a rated risk profile
                      & Materially faithful to input? & 3,5     & 0.08 & Pass \\
$s_4$ Referral packet & Render the underwriter-facing file
                      & Materially faithful to input? & 1       & 0.07 & Pass \\
$s_5$ Review          & Underwriter reads the packet and approves
                      & Authorized person reviewed?   & 1       & ---  & Pass \\
\midrule
Bind                  & Commit on the underwriter's approval
                      & ---                           & 1       & 0.19 & Fail \\
\bottomrule
\end{tabularx}
\caption{Illustrative context collapse workflow. Each hop stays under
the materiality tolerance, but the drift accumulates to 0.19 relative to the
submission, flipping the decision from refer to accept.}
\label{tab:context-collapse}
\end{table*}

Context Collapse is a compositional policy violation of a different shape. A decision is committed only after an authorised person reviews the file; every hand-off along the way is a faithful summary of what it received; and yet the file the reviewer sees supports a different decision from the one the original submission supports. No summary lies. Each drops only a little, and each drop is defensible on its own - but the drops accumulate down the chain, and by the end enough has been lost to flip the call. The review was real. It was performed on a file that no longer said what the submission said.
 
Consider referral review in commercial underwriting, shown in Table~\ref{tab:context-collapse}. The carrier's policy is that any account flagged for referral must receive substantive underwriter review before binding. Submissions arrive as heterogeneous packages - broker email, ACORD forms, loss runs, inspection reports - that no downstream stage reads in full, so each stage condenses what it received for the next.

Intake records five facts that matter to the decision: an ambiguous class code
carrying materially different rates, an unverified sprinkler certificate, a prior
large loss marked under investigation, a coverage gap between prior carriers, and
a broker note reporting pending litigation. Read whole, the submission supports
referral. Enrichment merges the file into a consolidated narrative and drops the
sprinkler certificate, since third-party data now supplies a protection class.
Risk profiling condenses the file to a rated profile: the coverage gap and the
pending litigation go, neither mapping to a rating factor, and the prior loss
survives as a headline figure, \$180K, with the qualifier ``under
investigation'' removed. Every word that remains is true and the number is exact,
but a loss and a loss under investigation mean different things to an
underwriter. The referral packet renders the underwriter-facing file, keeping the
ambiguous class code. The underwriter reads a packet carrying one of the five
facts and approves. The account binds on a decision of accept, when the
submission supported refer.

Nowhere did a component fail. Intake extracted accurately, each summary was a faithful condensation of its input, and the underwriter read the packet carefully and reached the decision it supported. No fact was fabricated, no step skipped, no authority exceeded. The account still bound on the wrong decision.

This pattern, the fact kept and the qualifier that told you how to read it
removed, is what Lee et al.~\cite{lee2026summaries} term \emph{decontextualisation}; they
find that under a fixed budget the share of such context-setting facts falls from
25\% in the source to 9\% after a single compression.

\paragraph{Why per-step monitoring is blind to it.} The failure is invisible to per-step monitoring, and invisible by design. Ask what each summary's checker can compare against. It can only compare its summary to the file it was handed - the previous summary. It cannot compare against the original submission, because the submission is gone by then; discarding it is the whole point of summarising. So every checker verifies faithfulness to the previous step, and no checker verifies faithfulness to the source. A chain of individually faithful hand-offs is exactly what the violation is made of, so no amount of accuracy added to the individual checkers can catch it. No step-level monitor holds both the source submission and the file the reviewer sees. 
 
Threshold Laundering and Context Collapse have opposite temporal blind spots. In Threshold Laundering, an earlier gate cannot observe policy-relevant changes that occur later in the workflow. In Context Collapse, the final decision-maker cannot recover policy-relevant information that was present in the original submission but lost through intermediate transformations.
 
A natural objection is that the summaries are simply too aggressive - give them more room. But the loss does not fall away with a larger budget: decision flips barely move even as the summary budget grows~\cite{lee2026summaries}. Bigger summaries drift more gently, but they still flip the decision. It is also not a failure of the reviewer, and not simply over-trust in the machine. Over-trust in automated output is a disposition that training or incentives can address~\cite{goddard2012automation}; this is not that. Even a maximally skeptical, perfectly sharp underwriter reaches the same decision, because the facts they would need are not on the page. You cannot scrutinize what is no longer there.
 
Then why not require each summary to preserve the decision exactly, rather than just closely? Because that check is one nobody can run. To verify that a summary preserves the same decision-relevant meaning as its input, the monitor would need to determine what decision would be reached from the full pre-summary information and compare it with the decision reached from the summary alone. In effect, this requires re-evaluating the account twice at every hand-off. A workflow able to do that at each step would not need the downstream steps or the reviewer at all. What is checked in practice is cheaper and narrower - that a summary is factually accurate and its claims trace to its input - and none of that constrains how far the decision has drifted. This is why the only real remedy is to keep the original submission and re-derive the decision from it before binding, which is a check over the whole trace rather than any single step.

\section{Repair Topology is Dictated by Violation Structure}
\label{sec:gateplacement}

The correct number and placement of policy gates is not a design choice. It is dictated by where the guarded quantity is mutated over the course of the workflow, and that differs by class.

\begin{figure}[H]
  \centering
  \includegraphics[width=1.0\columnwidth]{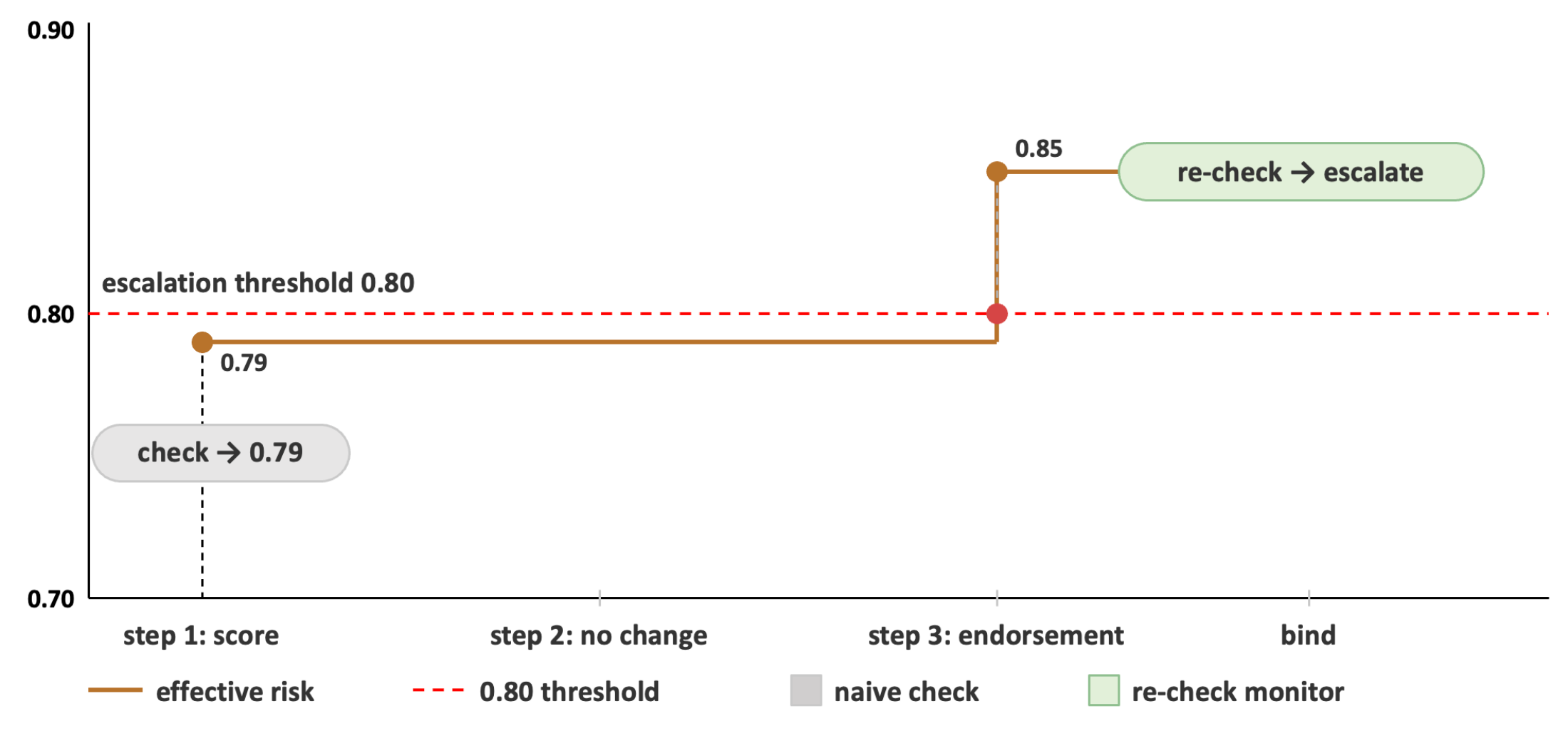}
  \caption{Gate Placement for Threshold Laundering}
  \label{fig:gateplacement}
\end{figure}

Consider threshold laundering, shown in figure~\ref{fig:gateplacement}: risk is not a static property but a quantity that mutates across steps. A single gate at step 1 sees $\mathrm{risk} = 0.79$, passes, and implicitly assumes that value will not change. Step 3 adds exposure the gate never observes, and the value crosses the threshold $0.80$ in a place where nobody was looking. The gate enforced a point-in-time predicate, $\mathrm{risk} < 0.80$ at step 1, when the governing policy specifies a invariant: $\mathrm{risk} < 0.80$ on the committed state.

The same principle governs the other three classes, though what mutates differs. In authority creep, what mutates is not a scalar but the representation of who is entitled to decide: the escalation router already sits at the only step where escalation authority is exercised, so there is no later position to move it to. What must be added is not a new position but backward reach. The router must be able to trace the count it fires on through the domain-scoped transforms that produced it, since the mutation happened upstream of any single gate rather than after one. In Cumulative Sum Violation(CSV), the mutation is a running total that accretes across independently-gated actions, and no gate exists at the scope where that total is tracked at all; placement is not wrong here so much as absent, and the fix is to instantiate a gate at the aggregate's own scope rather than to relocate an existing one. In Context Collapse Violation, what mutates is the evidentiary record itself: each hand-off is a faithful summary of what it received, so the terminal review gate is correctly placed and correctly applied, but the file it evaluates is no longer the file the policy was written against. The mutation is in what the gate can see, not in the guarded quantity or in where the gate sits.

Each class therefore demands a different repair topology:
\begin{itemize}
    \item \textbf{Authority Creep:} Trace backward once from the decision point.
    \item \textbf{Threshold Laundering:} Re-evaluate the existing predicate on the committed state rather than the value present when the gate first fired.
    \item \textbf{Cumulative Sum Violation:} Introduce a predicate at the aggregate's own scope, since no step currently holds one.
    \item \textbf{Context Collapse:} Reconstruct against the original submission rather than the intermediate summaries.
\end{itemize}

This is why the detector cannot be a collection of single-point checkpoints. It must reconstruct full policy semantics and re-evaluate against the final committed state, which only the complete trace provides.

\section{CPV Detection Architecture}
\label{sec:architecture}

The repair topologies established in Section~\ref{sec:gateplacement} cannot be implemented by a single monolithic check. Each demands a different kind of infrastructure: a persistent, unabridged record of the execution to trace or monitor against, and the ability to recompute a guarded quantity from that record rather than trust an intermediate representation a prior step may have already laundered. The architecture therefore separates the concern of preserving the trace from the concern of evaluating policy against it. The detector comprises four core stages:

\begin{enumerate}[leftmargin=1.4em]
    \item \textbf{Provenance Ingestion.} Collect and normalize runtime events into a canonical, timestamped trace.
    \item \textbf{State Reconstruction.} Rebuild workflow state and policy-relevant history from raw provenance.
    \item \textbf{Policy Evaluation.} Evaluate both step-level and workflow-level policies against reconstructed state.
    \item \textbf{Compositional Detection.} Identify violations that emerge only at the sequence level and classify by CPV type.
\end{enumerate}

Two design invariants govern all four:
\begin{itemize}[leftmargin=1.4em]
    \item \textbf{History-completeness.} No component may render a violation verdict from a truncated or windowed view when the governing policies semantics require the full trajectory.
    \item \textbf{Recount-from-raw-provenance.} Composition gates must recompute guarded quantities from raw provenance under the policies own semantic definition, never from the pipeline's transformed or derived representation.
\end{itemize}

\begin{figure}[H]
  \centering
  \includegraphics[width=1.0\columnwidth]{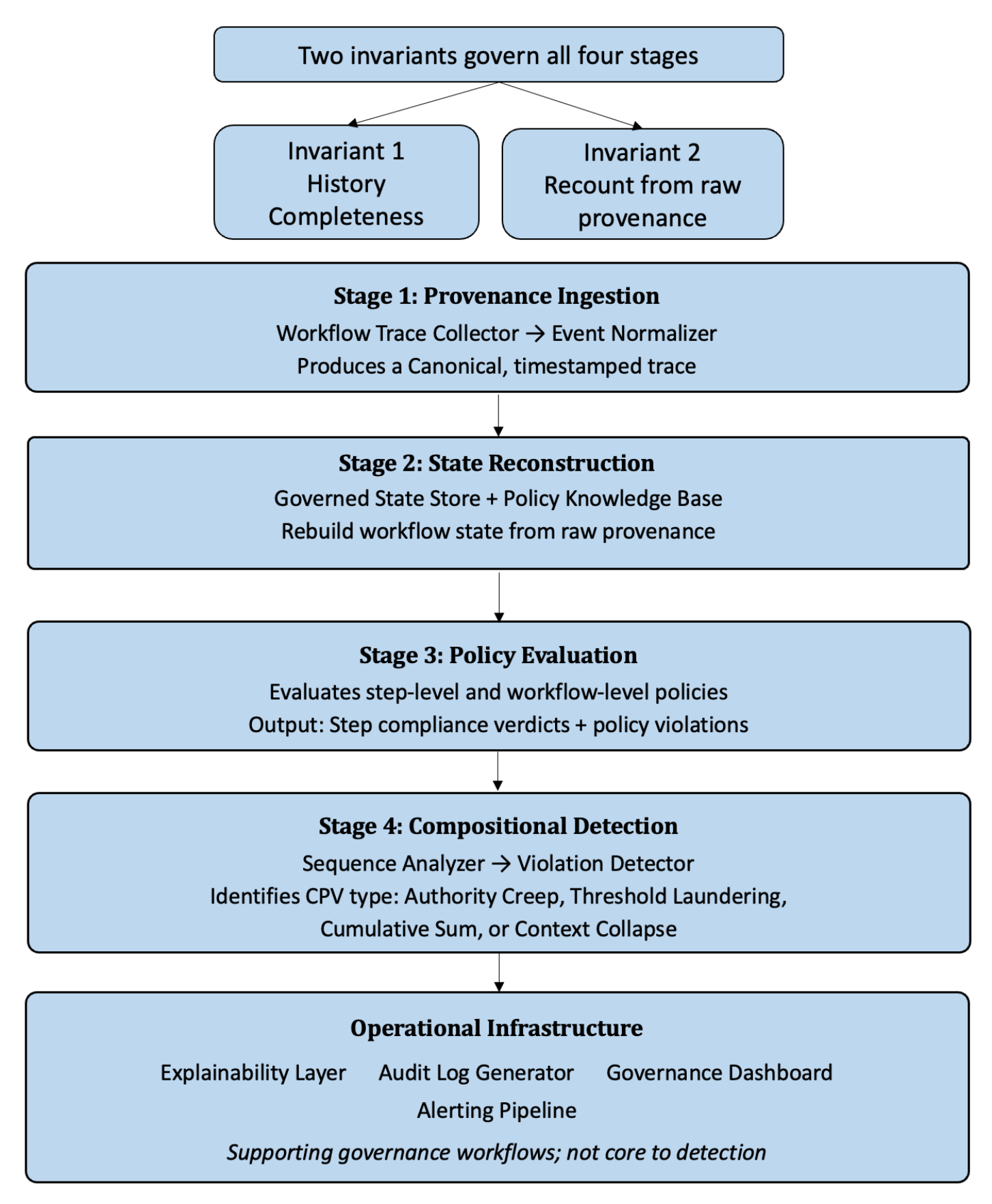}
  \caption{CPV Detector architecture. Four detection stages, governed throughout by
two design invariants, with operational components that support governance
workflows but do not contribute to detection logic}
  \label{fig:cpv-detector}
\end{figure}

\subsection{Stage 1: Provenance Ingestion and Normalization}
\begin{itemize}[leftmargin=1em]
\item \textbf{Workflow Trace Collector:} Captures every agent-relevant event, including prompts, tool calls, approvals, memory writes, external API responses, file access, retries, and handoffs. It ingests raw runtime events, tool telemetry, orchestrator logs, and identity context and produces an immutable, ordered event log for each workflow instance with monotonic sequence numbers and timestamps.

\item \textbf{Event Normalizer:} Converts heterogeneous runtime events into a unified workflow-trace schema while preserving raw provenance references. It consumes raw events from collectors and produces canonical events containing instance identifiers, sequence numbers, timestamps, actors, action types, typed parameters, and provenance links. Internally, it maps framework-specific artifacts into standardized event categories and enriches them with semantic attributes such as action class, resource class, sensitivity level, and purpose.

These two components establish the complete, auditable record required by history-completeness. Nothing is discarded; later stages see the full execution history.
\end{itemize}
\subsection{Stage 2: State Reconstruction}
\begin{itemize}[leftmargin=1em]
\item\textbf{Governed State Store:} Persists workflow state and the raw provenance ledger for each workflow instance, enabling recount-from-raw analysis. It ingests raw payloads from the Workflow Trace Collector, canonical events from the Event Normalizer, and derived annotations from downstream engines, providing queryable current and historical state with provenance retrieval by instance ID.

\item\textbf{Policy Knowledge Base (PKB):} Stores machine-readable policies, control objectives, exceptions, and versioned policy definitions, serving as the authoritative source for policy semantics required by the recount principle.
\end{itemize}
These components implement the recount-from-raw-provenance principle: the State Store holds the raw facts, the PKB holds the policy semantics, and together they enable the detector to reconstruct what the policy actually requires at each decision point.

\subsection{Stage 3: Policy Evaluation}
\begin{itemize}[leftmargin=1em]
\item\textbf{Policy Evaluation:} Evaluates declarative policies, authority constraints, and separation-of-duty rules against the workflow trace and governed state. It produces policy-level pass/fail decisions with supporting facts.

This engine answers two questions:
\begin{enumerate}[leftmargin=1.4em,itemsep=1pt,topsep=2pt]
  \item Do all individual steps satisfy their local policies?
        (step-level compliance)
  \item Does the composed workflow satisfy trace-level policies?
        (workflow-level compliance)
\end{enumerate}

The output is the raw material for compositional detection: a record of which
steps are locally compliant, and which policies defined over the complete trace
are violated.
\end{itemize}
\subsection{Stage 4: Compositional Detection}
\begin{itemize}[leftmargin=1em]
\item\textbf{Sequence Analyzer:} This layer analyzes complete workflow trajectories to identify compositional policy violations that emerge only across multiple steps.

The Sequence Analyzer performs the core detection logic, and the check it runs
follows the repair topology of each class:
\begin{itemize}[leftmargin=1.4em,itemsep=1pt,topsep=2pt]
  \item \textbf{Authority Creep.} Traces authority ownership backward from the
        decision gate to determine whether accumulated permissions govern every
        input the gate depends on.
  \item \textbf{Threshold Laundering.} Re-evaluates the guarded quantity on the
        committed state and compares it against the value present when the gate
        originally fired.
  \item \textbf{Cumulative Sum Violation.} Maintains the aggregate over the
        trace and compares it against the policy threshold no step holds.
  \item \textbf{Context Collapse.} Re-derives the decision from the retained
        submission and compares it against the decision the reviewed file
        supports.
\end{itemize}

\item\textbf{Violation Detector:} Produces authoritative binary or graded violation verdicts for each workflow instance and maps detected issues to the CPV taxonomy with violation type, severity, and the responsible workflow segment. Confirmed findings are classified into known CPV categories, while unmatched patterns are flagged as candidates for further analysis.
\end{itemize}

\subsection{Operational Components}
The detector also includes components supporting deployment and governance workflows:
\begin{itemize}[leftmargin=1em]
\item\textbf{Explainability Layer:} Generates human-auditable justifications that trace each violation verdict to the responsible workflow events, raw provenance records, and policy definitions. This layer provides explanations to the Governance Dashboard and Audit Log Generator.

\item\textbf{Audit Log Generator:} Produces tamper-evident, regulator-grade records of violation verdicts, explanations, and supporting evidence. The audit log serves as a durable sink for records from upstream components.

\item\textbf{Governance Dashboard \& Alerting Pipeline:} Provides portfolio- and instance-level visibility into CPV posture and routes actionable CPV findings to human reviewers and enforcement systems. 
\end{itemize}

These are operational infrastructure supporting the core detection logic; they do not contribute to CPV identification itself.

The CPV Detector architecture operationalizes trace-level governance through four core detection stages: Provenance Ingestion normalizes runtime events into a canonical trace; State Reconstruction maintains complete history and raw provenance; Policy Evaluation checks both step-level and workflow-level policies; and Compositional Detection identifies violations that emerge only at the sequence level. This linear flow is anchored by two foundational design invariants: history-completeness ensures no violation verdict is rendered from a truncated view, and recount-from-raw-provenance ensures all policy-relevant quantities are recomputed from original provenance, not intermediate pipeline representations. These invariants are not optional refinements - they are structural necessities. This is why the architecture cannot collapse into step-level checks; it must work at the workflow level, treating policy enforcement as a flow-level invariant rather than a post-hoc monitoring layer.

\subsection{Detector Failure Modes}

The architecture inherits limits from the trace it is given, and we state
them explicitly.

\textbf{Provenance loss:} The detector can only recompute a guarded
quantity from what the pipeline retained. If a transform deletes suppressed
evidence rather than marking it, the pre-image is unrecoverable and the
violation is undetectable in principle - not hard, but impossible, since no
information remains in the trace from which the truth is derivable.
Provenance retention is therefore a precondition rather
than an implementation detail, and a pipeline that discards what it
suppresses cannot be audited for Authority Creep regardless of what is
added downstream.

\textbf{Oracle dependence for subjective predicates:} Recomputation is sound when the guarded quantity is objective. Many predicates are not: whether an exception is material applies a standard written in prose. The detector then requires an interpretation oracle, which becomes part of its trusted base. If that oracle and the pipeline component that produced the violation are both language models reasoning from similar priors, they may reach the same judgement; the detector sees no discrepancy and concurs. A recount is sound only where its oracle is independent of the pipeline's and anchored to the policy text rather than to learned precedent.

\textbf{Extraction error:} Even with full provenance, recomputation from documents inherits extraction error: boundary-band dates misclassified against a lookback window, the same loss double-counted under distinct claim numbers, differing totals read from different documents. Trace-level evaluation removes a structural blindness; it does not remove measurement error.

\textbf{Partial and asynchronous observability:} The architecture assumes
an ordered event log per workflow instance. Distributed execution, clock
skew, out-of-order delivery, and external tools that do not emit telemetry
all degrade this assumption, and runtime verification under incomplete or
distributed observation is known to admit inconclusive
verdicts~\cite{basin2015failureaware,bauer2011runtime}.

\textbf{Entity resolution:} Cumulative Sum Violations are defined over the
aggregate belonging to an entity. Fragmented identities, aliases, or
multiple accounts prevent construction of the correct aggregate, and no
amount of trace retention repairs an aggregate assembled over the wrong
partition.

\textbf{Hidden state:} Agent memory, retrieved context, and reasoning not
surfaced as events lie outside the trace. Where policy-relevant state is
carried in such channels, the reconstructed trajectory is incomplete and
the detector's verdict is correspondingly weaker.

\textbf{Cost:} Retaining raw provenance and recomputing guarded quantities
at commit time imposes storage and latency costs that scale with trace
length and with the number of gated quantities; for Context Collapse in
particular, re-deriving the decision from the source submission approaches
the cost of the un-automated decision itself.

\section{Future Research}

As governance systems scale to handle compositional violations, key challenges emerge around deployment, generalization, and trust. Operationalizing the detector introduces engineering considerations, provenance collection overhead, persistent event log storage, latency in real-time policy evaluation, and efficient state reconstruction in distributed workflows. These are important implementation concerns but are orthogonal to the core detection mechanism. Production deployments will need to make storage-latency-accuracy trade-offs specific to workflow volume and SLA requirements, leveraging tiered storage, batch vs. streaming policy evaluation, and incremental state reconstruction.

Beyond implementation, standardized frameworks for expressing governance logic and mapping it to workflow provenance remain absent. Whether violation mechanisms and detector architecture transfer to lending, hiring, moderation, and other high-stakes domains is unclear. Remediation strategies, accountability assignment, and defenses against adversarial obfuscation of audit trails raise practical and ethical questions beyond detection itself.

Future research should focus on: (i) standardized policy languages and semantic mappings between governance intent and workflow provenance, enabling transparent detection across domains; (ii) simulation testbeds and benchmarked datasets analogous to those in fairness and interpretability to lower barriers to entry and enable performance comparison; (iii) efficient trajectory analysis and real-time provenance capture to bring detection costs within operational feasibility; and (iv) domain-specific instantiation and validation in regulated workflows

\section{Conclusion}
We have identified compositional policy violations as a distinct and consequential class of governance failure: sequences where every individual step is locally policy-compliant yet the composed trajectory violates organizational governance, regulatory obligations, or business intent. By formalizing four violation mechanisms, deriving the repair topology each one demands, and proposing a detector architecture that describes on how to reconstruct full policy semantics from raw provenance, we have established that point-in-time checkpoint enforcement is structurally insufficient. This work opens a research agenda in policy-aware AI systems design - one that treats policy enforcement not as an external checkpoint but as a standing property of the workflow, holding across the full trace rather than at isolated decision points. As agentic AI workflows and production AI systems grow in autonomy and composition, the ability to detect violations that emerge from sequences of individually compliant steps becomes not merely an engineering concern but a prerequisite for organizational accountability and regulatory compliance. Compositional violations are not edge cases; they are inherent to how autonomous agents compose decisions at scale, and detecting them is foundational to trustworthy deployment.

\bibliographystyle{unsrt}  
\bibliography{references} 

@inproceedings{ahad2026semantic,
  title={Semantic Intent Fragmentation: A Single-Shot Compositional Attack on Multi-Agent AI Pipelines},
  author={Ahad, Tanzim and Hossain, Ismail and Alam, Md Jahangir and Puppala, Sai and Lee, Yoonpyo and Alam, Syed Bahauddin and Talukder, Sajedul},
  booktitle={Proceedings of the AAAI Symposium Series},
  volume={9},

  pages={229--237},
  year={2026}
}

@article{lee2026summaries,
  title={When Summaries Distort Decisions: Information Fidelity in LLM-Compressed Financial Analysis},
  author={Lee, Hoyoung and Park, Suhwan and Lee, Seunghan and Seo, Jun and Lee, Jaehoon and Yoo, Sungdong and Kim, Minjae and Na, CheolWon and Wang, Zhangyang and Golkhou, Zach and others},
  journal={arXiv preprint arXiv:2606.29251},
  year={2026}
}

@misc{uscode5324,
  title        = {31 U.S.C. § 5324: Structuring Transactions to Evade Reporting Requirements},
  author       = {{United States Code}},
  year         = {2024},
  note         = {Bank Secrecy Act provisions regarding transaction structuring}
}

@inproceedings{basin2015failureaware,
  author    = {Basin, David and Klaedtke, Felix and Z{\v a}linescu, Eugen},
  title     = {Failure-aware Runtime Verification of Distributed Systems},
  booktitle = {Proceedings of the 35th IARCS Annual Conference on Foundations 
               of Software Technology and Theoretical Computer Science},
  year      = {2015}
}

@article{bauer2011runtime,
  title={Runtime verification for LTL and TLTL},
  author={Bauer, Andreas and Leucker, Martin and Schallhart, Christian},
  journal={ACM Transactions on Software Engineering and Methodology (TOSEM)},
  volume={20},
  number={4},
  pages={1--64},
  year={2011},
  publisher={ACM New York, NY, USA}
}

@article{wang2025agentspec,
  title={Agentspec: Customizable runtime enforcement for safe and reliable llm agents},
  author={Wang, Haoyu and Poskitt, Christopher M and Sun, Jun},
  journal={arXiv preprint arXiv:2503.18666},
  year={2025}
}

@article{goddard2012automation,
  title={Automation bias: a systematic review of frequency, effect mediators, and mitigators},
  author={Goddard, Kate and Roudsari, Abdul and Wyatt, Jeremy C},
  journal={Journal of the American Medical Informatics Association},
  volume={19},
  number={1},
  pages={121--127},
  year={2012},
  publisher={BMJ Group BMA House, Tavistock Square, London, WC1H 9JR}
}

@misc{anthropic2025financial,
  title        = {Claude for Financial Services},
  author       = {{Anthropic}},
  year         = {2025},
  howpublished = {\url{https://www.anthropic.com/news/claude-for-financial-services}},
  note         = {Official announcement; contains AIG CEO statement}
}

@misc{allianz2026anthropic,
  title        = {Allianz and Anthropic Forge Global Partnership to Advance
                  Responsible {AI} in Insurance},
  author       = {{Allianz}},
  year         = {2026},
  howpublished = {\url{https://www.allianz.com/en/mediacenter/news/media-releases/260109-allianz-and-anthropic-forge-global-partnership.html}},
  note         = {Official press release}
}

@misc{baldwin2026anthropic,
  title        = {The Baldwin Group Announces Expanded Enterprise Relationship
                  with Anthropic},
  author       = {{The Baldwin Group}},
  year         = {2026},
  howpublished = {\url{https://ir.baldwin.com/news-releases/news-release-details/baldwin-group-announces-expanded-enterprise-relationship/}},
  note         = {Official press release}
}

@article{inan2023llamaguard,
  title   = {Llama Guard: {LLM}-based Input-Output Safeguard for Human-{AI}
             Conversations},
  author  = {Inan, Hakan and Upasani, Kartikeya and Chi, Jianfeng and Rungta,
             Rashi and Iyer, Krithika and Mao, Yuning and Tontchev, Michael and
             Hu, Qing and Fuller, Brian and Testuggine, Davide and Khabsa, Madian},
  journal = {arXiv preprint arXiv:2312.06674},
  year    = {2023},
  doi     = {10.48550/arXiv.2312.06674},
  url     = {https://arxiv.org/abs/2312.06674},
  note    = {Preprint}
}

@inproceedings{rebedea2023nemo,
  title     = {{NeMo Guardrails}: A Toolkit for Controllable and Safe {LLM}
               Applications with Programmable Rails},
  author    = {Rebedea, Traian and Dinu, Razvan and Sreedhar, Makesh Narsimhan
               and Parisien, Christopher and Cohen, Jonathan},
  booktitle = {Proceedings of the 2023 Conference on Empirical Methods in
               Natural Language Processing: System Demonstrations},
  pages     = {431--445},
  year      = {2023},
  doi       = {10.18653/v1/2023.emnlp-demo.40},
  url       = {https://aclanthology.org/2023.emnlp-demo.40/},
  note      = {Peer-reviewed}
}

@article{sharma2025constitutional,
  title   = {Constitutional Classifiers: Defending against Universal Jailbreaks
             across Thousands of Hours of Red Teaming},
  author  = {Sharma, Mrinank and Tong, Meg and Mu, Jesse and Wei, Jerry and
             Kruthoff, Jorrit and Goodfriend, Scott and Ong, Euan and Peng, Alwin
             and Agarwal, Raj and Anil, Cem and Perez, Ethan},
  journal = {arXiv preprint arXiv:2501.18837},
  year    = {2025},
  url     = {https://arxiv.org/abs/2501.18837},
  note    = {Preprint (Anthropic)}
}

@misc{langfuse2026datamodel,
  title        = {Observability Data Model},
  author       = {{Langfuse}},
  howpublished = {\url{https://langfuse.com/docs/observability/data-model}},
  year         = {2026},
  note         = {Official documentation; accessed 2026}
}

@misc{langsmith2026concepts,
  title        = {Observability Concepts},
  author       = {{LangChain}},
  howpublished = {\url{https://docs.langchain.com/langsmith/observability-concepts}},
  year         = {2026},
  note         = {Official documentation; accessed 2026}
}

@article{azarafrooz2026crosssession,
  title   = {Cross-Session Threats in {AI} Agents: Benchmark, Evaluation, and
             Algorithms},
  author  = {Azarafrooz, Mahdi},
  journal = {arXiv preprint arXiv:2604.21131},
  year    = {2026},
  url     = {https://arxiv.org/abs/2604.21131},
  note    = {Preprint}
}

@article{dhodapkar2026safetydrift,
  title   = {{SafetyDrift}: Predicting When {AI} Agents Cross the Line Before
             They Actually Do},
  author  = {Dhodapkar, Aniruddha and Pishori, Fatima},
  journal = {arXiv preprint arXiv:2603.27148},
  year    = {2026},
  url     = {https://arxiv.org/abs/2603.27148},
  note    = {Preprint}
}

@inproceedings{mireshghallah2026cimemories,
  title     = {{CIMemories}: A Compositional Benchmark for Contextual Integrity
               of Persistent Memory in {LLM}s},
  author    = {Mireshghallah, Niloofar and Mangaokar, Neal and Kokhlikyan, Narine
               and Zharmagambetov, Arman and Zaheer, Manzil and Mahloujifar, Saeed
               and Chaudhuri, Kamalika},
  booktitle = {International Conference on Learning Representations (ICLR)},
  year      = {2026},
  url       = {https://arxiv.org/abs/2511.14937},
  note      = {Peer-reviewed}
}

@inproceedings{lightman2024verify,
  title     = {Let's Verify Step by Step},
  author    = {Lightman, Hunter and Kosaraju, Vineet and Burda, Yura and
               Edwards, Harri and Baker, Bowen and Lee, Teddy and Leike, Jan and
               Schulman, John and Sutskever, Ilya and Cobbe, Karl},
  booktitle = {International Conference on Learning Representations (ICLR)},
  year      = {2024},
  url       = {https://arxiv.org/abs/2305.20050},
  note      = {Peer-reviewed}
}

@inproceedings{cemri2025mast,
  title     = {Why Do Multi-Agent {LLM} Systems Fail?},
  author    = {Cemri, Mert and Pan, Melissa Z. and Yang, Shuyi and Agrawal,
               Lakshya A. and Chopra, Bhavya and Tiwari, Rishabh and Keutzer, Kurt
               and Parameswaran, Aditya and Klein, Dan and Ramchandran, Kannan and
               Zaharia, Matei and Gonzalez, Joseph E. and Stoica, Ion},
  booktitle = {Advances in Neural Information Processing Systems (NeurIPS)},
  year      = {2025},
  url       = {https://arxiv.org/abs/2503.13657},
  note      = {Peer-reviewed}
}

@article{green2022oversight,
  title   = {The Flaws of Policies Requiring Human Oversight of Government
             Algorithms},
  author  = {Green, Ben},
  journal = {Computer Law \& Security Review},
  volume  = {45},
  pages   = {105681},
  year    = {2022},
  doi     = {10.1016/j.clsr.2022.105681},
  url     = {https://www.sciencedirect.com/science/article/pii/S0267364922000292},
  note    = {Peer-reviewed}
}
\end{document}